\documentclass[11pt]{article}

\usepackage[preprint]{acl}

\usepackage{times}
\usepackage{latexsym}

\usepackage[T1]{fontenc}

\usepackage[utf8]{inputenc}

\usepackage{microtype}

\usepackage{inconsolata}
\usepackage{booktabs}
\usepackage{multirow}
\usepackage{tabularx}
\usepackage{listings}
\usepackage[many]{tcolorbox}
\usepackage{longtable}
\usepackage{xcolor}
\usepackage{dcolumn}
\usepackage{siunitx}
\usepackage{subcaption}
\usepackage[T1]{fontenc}
\newcolumntype{R}{>{\raggedleft\arraybackslash}X}
\usepackage{graphicx}

\title{Automated Analysis of Sustainability Reports: Using Large Language Models for the Extraction and Prediction of EU
Taxonomy-Compliant KPIs}

\author{Jonathan Schmoll \\
  \texttt{jonathan.schmoll@student.uibk.ac.at} \\ University of Innsbruck \\\And
 Adam Jatowt \\
  \texttt{adam.jatowt@uibk.ac.at} \\ University of Innsbruck  }

\begin{document}
\maketitle
\begin{abstract}
The manual, resource-intensive process of complying with the EU Taxonomy presents a significant challenge for companies. While Large Language Models (LLMs) offer a path to automation, research is hindered by a lack of public benchmark datasets. To address this gap, we introduce a novel, structured dataset from 190 corporate reports, containing ground-truth economic activities and quantitative Key Performance Indicators (KPIs). We use this dataset to conduct the first systematic evaluation of LLMs on the core compliance workflow.
Our results reveal a clear performance gap between qualitative and quantitative tasks. LLMs show moderate success in the qualitative task of identifying economic activities, with a multi-step agentic framework modestly enhancing precision. Conversely, the models comprehensively fail at the quantitative task of predicting financial KPIs in a zero-shot setting. We also discover a paradox, where concise metadata often yields superior performance to full, unstructured reports, and find that model confidence scores are poorly calibrated. We conclude that while LLMs are not ready for full automation, they can serve as powerful assistive tools for human experts. Our dataset provides a public benchmark for future research.
\end{abstract}

\section{Introduction}
\label{sec:introduction}

The European Union (EU) is mobilizing significant capital towards sustainable finance, guided by the European Green Deal \cite{ec:ip_20_17}. A cornerstone of this strategy is the EU Taxonomy \cite{eu:taxonomyRegulation2020}, a complex classification system designed to provide a standard, science-based definition of sustainable economic activities. The regulation's primary goal is to create market transparency, combat greenwashing, and direct investment toward activities essential for the green transition \cite{ec:taxonomy_overview,SNEIDERIENE2025102720,calamai2025corporategreenwashingdetectiontext}.

This framework is operationalized through the Corporate Sustainability Reporting Directive (CSRD) \cite{eu:csrd:2022}, which legally mandates that thousands of companies analyze their operations and publicly disclose their alignment with the Taxonomy. This compliance process requires undertakings to perform a complex, manual analysis of dense legal acts and unstructured internal documents to execute two core information extraction (IE) tasks: (1) identifying which of the many official economic activities the company performs, and (2) extracting the precise financial Key Performance Indicators (KPIs), Turnover, CapEx, and OpEx, associated with those activities \cite{eu:disclosuresDelegatedAct2021}. This manual workflow is resource-intensive, prone to misinterpretation, and scales poorly as the regulation evolves \cite{eu:climateDelegatedAct2021, eu:environmentalDelegatedAct2023,schuetze2025}.

The recent advancements in Large Language Models (LLMs) for complex reasoning and IE present a compelling opportunity to automate or assist these analytical tasks. However, the practical viability of general-purpose LLMs in this specialized, high-stakes domain is unproven. A significant barrier to research and development in this area is the lack of public, structured benchmark datasets; corporate disclosures are typically locked in unstructured PDF reports, making it difficult to evaluate or train models for these specific IE tasks.

To address this gap, this paper presents two primary contributions. First, we introduce and make publicly available a \textbf{novel, structured benchmark dataset} for EU Taxonomy analysis, curated from the annual reports of 190 companies. For each company, the dataset provides machine-readable reports, verified ground-truth lists of declared economic activities, and the corresponding quantitative KPIs (Turnover, CapEx, OpEx). Second, we use this dataset to conduct the \textbf{first systematic evaluation of modern LLMs} on the core compliance workflow, which we formulate as four distinct tasks: (1) multiclass activity prediction, (2) binary activity classification, (3) quantitative KPI regression, and (4) a multi-step agentic workflow.

Our empirical results reveal a clear performance gap between qualitative and quantitative tasks. We find that LLMs demonstrate moderate success in the qualitative task of identifying relevant economic activities, with a multi-step agentic framework modestly enhancing precision. Conversely, the models comprehensively fail at the quantitative task of predicting financial KPIs in a zero-shot setting, yielding results no better than a baseline mean prediction. We also discover a "paradox of context," where concise, structured metadata often yields superior classification performance compared to the full, unstructured text of an annual report. Finally, we show that the models' self-reported confidence scores are poorly calibrated and thus unreliable for this task.

We conclude that while current-generation LLMs are not yet suitable for fully automated, end-to-end compliance, they can serve as powerful assistive tools to accelerate the discovery and analysis phases for human experts. Our dataset and baseline results provide a public benchmark for future research into developing specialized models for this complex and growing regulatory domain.

\section{The EU Taxonomy Reporting Task}
\label{sec:taxonomy_task}

\begin{figure*}[t]
    \captionsetup{justification=centering}
    \centering
    \includegraphics[width=.7\textwidth]{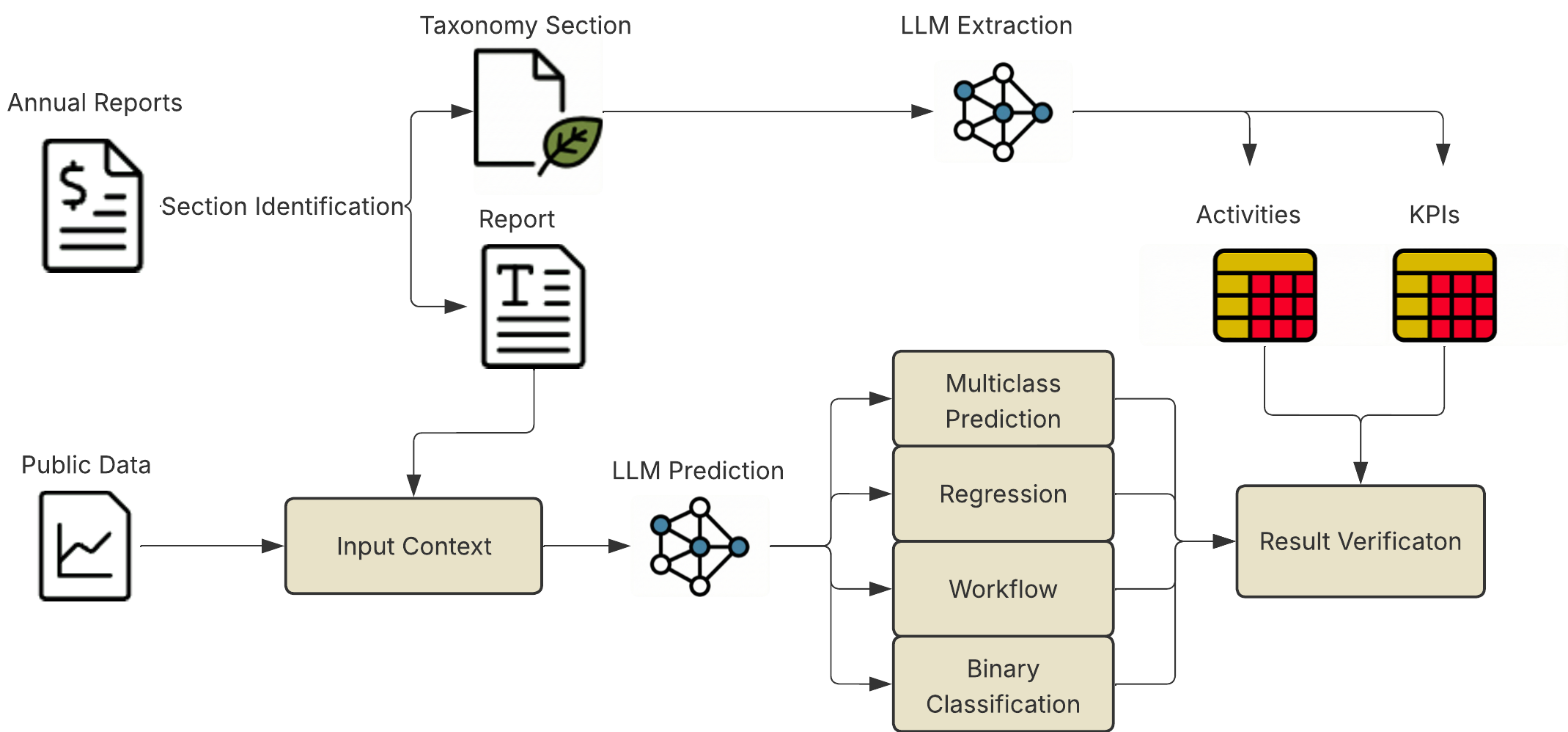}
    \caption{Overview: Extraction of source data from public reports. Multiple zero-shot experiments against the dataset to test current model capabilities.}
    \label{fig:overview_fig}
\end{figure*}

\subsection{Regulatory Context and Reporting Obligations}
The EU Taxonomy is a comprehensive classification system established by Regulation (EU) 2020/852 to define environmentally sustainable economic activities \cite{eu:taxonomyRegulation2020}. It serves as a key component of the European Green Deal's sustainable finance framework, aiming to increase market transparency and combat \emph{greenwashing} \cite{ec:taxonomy_overview, ec:green_deal_strategy}.

The regulation is operationalized through the \emph{Corporate Sustainability Reporting Directive (CSRD)} \cite{eu:csrd:2022}, which mandates that a wide range of undertakings publicly disclose their alignment with the Taxonomy. This compliance process requires companies to perform a complex internal assessment. First, an activity must make a \emph{substantial contribution} to one of six environmental objectives (e.g., climate change mitigation). Second, it must \emph{do no significant harm} (DNSH) to the other five objectives, while also meeting minimum social safeguards \cite{eu:taxonomyRegulation2020}.

The core output of this assessment is the disclosure of three mandatory Key Performance Indicators (KPIs), detailed in a specific \emph{Disclosures Delegated Act} \cite{eu:disclosuresDelegatedAct2021}. Non-financial undertakings must report the proportion of their operations that are Taxonomy-aligned, quantified across:
\begin{description}
    \item[Turnover KPI:] The proportion of net turnover derived from sustainable products or services.
    \item[CapEx KPI:] The proportion of capital expenditures associated with sustainable assets or processes.
    \item[OpEx KPI:] The proportion of operational expenditures related to sustainable activities.
\end{description}

\subsection{Formulating the IE Tasks: Activity Classification and KPI Regression}
\label{sec:formulating_tasks}
For undertakings, the reporting process is a resource-intensive, manual workflow that involves interpreting dense legal texts (\emph{Delegated Acts}) and mapping them to unstructured internal data from finance, legal, and sustainability departments \cite{eu:climateDelegatedAct2021, eu:environmentalDelegatedAct2023}. This manual process is error-prone, inconsistent, and scales poorly as the regulation evolves\cite{schuetze2025}.

This regulatory challenge can be framed as a complex information extraction (IE) and reasoning problem. We decompose this compliance workflow into two primary computational tasks, which form the basis of our evaluation:

\begin{itemize}
    \item \textbf{Task 1: Activity Classification.} This is a qualitative task. Before any financial data can be aggregated, a company must first determine \emph{which} of its operations correspond to the official list of economic activities defined in the Taxonomy. We model this as a multi-label text classification problem, where the input is a company's operational description (e.g., from an annual report) and the output is a predicted set of relevant Taxonomy activities.
    
    \item \textbf{Task 2: Quantitative KPI Regression.} This is a quantitative task. Once relevant activities are identified, the associated financial metrics must be extracted and quantified. We model this as a regression problem, where the model must predict the numerical percentage for the Turnover, CapEx, and OpEx KPIs based on the textual content of a company's report.
\end{itemize}

The primary challenge lies in the model's ability to semantically bridge the gap between unstructured, narrative descriptions of a company's business and the highly structured, technical definitions of the legal framework.

\section{Related Work}
\label{sec:related_work}

\subsection{NLP for Financial and ESG Reporting}
The application of Natural Language Processing (NLP) in the financial domain has rapidly evolved from traditional text mining to complex reasoning tasks using Large Language Models (LLMs) \cite{li2024largelanguagemodelsfinance}. A key driver has been the development of domain-specific models. \emph{FinBERT}, an adaptation of BERT pre-trained on financial documents, demonstrated high performance on financial sentiment analysis \cite{birti2025optimizinglargelanguagemodels}. More recently, models like \emph{BloombergGPT} have shown state-of-the-art performance on financial NLP benchmarks by training on a mix of general and domain-specific corpora \cite{li2024largelanguagemodelsfinance}.

This trend extends to the emerging field of Environmental, Social, and Governance (ESG) analysis. \emph{ESGReveal}, for instance, employs a Retrieval-Augmented Generation (RAG) approach to extract structured numerical and qualitative data from corporate ESG reports, with GPT-4 achieving high accuracy on disclosure analysis tasks \cite{zou2023esgrevealllmbasedapproachextracting}. Concurrently, benchmarks like \emph{ESGenius} have been developed to evaluate the foundational ESG knowledge of LLMs using expert-validated multiple-choice questions \cite{he2025esgeniusbenchmarkingllmsenvironmental}. While this work validates LLMs for general ESG analysis, our research focuses on the specific, practical IE tasks mandated by EU Taxonomy compliance, moving from broad knowledge assessment to direct application.

\subsection{Information Extraction from Regulatory Documents}
The core compliance tasks mandated by the EU Taxonomy, identifying relevant economic activities and extracting quantitative KPIs from corporate reports are applications of Information Extraction (IE). Recently, LLMs have catalyzed a shift in IE from traditional supervised, token-level classification (e.g., Named Entity Recognition or NER) to \emph{generative IE} \cite{xu2024largelanguagemodelsgenerative}. In this paradigm, extraction is reformulated as a text-to-text task, where a model is prompted to generate the desired structured output directly, often demonstrating strong zero-shot capabilities \cite{xu2024largelanguagemodelsgenerative}.

However, applying IE to specialized domains like legal and financial reporting presents distinct challenges, particularly the high cost of manual annotation \cite{Oliveira2025}. Research has shown that combining LLM-based prompting with weak supervision and a small fraction of human-labeled data can be effective for creating training data for NER in complex legal documents \cite{Oliveira2025}. More specific to our focus, \cite{birti2025optimizinglargelanguagemodels} introduced \emph{ESG-Activities}, a benchmark for classifying financial texts against EU Taxonomy activities. Their work highlights that fine-tuning smaller models on domain-specific data (both manual and synthetic) can significantly improve performance, even surpassing larger proprietary models \cite{birti2025optimizinglargelanguagemodels}.

Our research builds on these findings by addressing a key gap: prior studies have focused either on general ESG analysis \cite{zou2023esgrevealllmbasedapproachextracting, he2025esgeniusbenchmarkingllmsenvironmental} or a single sub-task of Taxonomy compliance (activity classification) \cite{birti2025optimizinglargelanguagemodels}. This paper presents the first systematic, multi-task evaluation of LLMs on the \emph{combined} workflow of both qualitative activity classification and quantitative KPI regression, using a new, comprehensive benchmark dataset derived from complete corporate reports.

\section{A Benchmark Dataset for EU Taxonomy Analysis}
\label{sec:dataset}
A significant barrier to research in automated EU Taxonomy compliance is the lack of public, structured benchmark datasets; corporate disclosures are typically locked in unstructured PDF reports. To address this, we introduce and make publicly available\footnote{Dataset available at \url{https://github.com/anonymous-duckling/acl}} a novel, structured dataset curated from the 2023 fiscal year reports of 190 companies subject to EU Taxonomy disclosure obligations.

\subsection{Dataset Curation and Preprocessing}
We collected 190 annual and sustainability reports from publicly listed companies, primarily from European stock exchanges (e.g., Frankfurter Börse, Wiener Börse) and corporate websites. All documents were originally in PDF format.

The curation process followed a multi-stage pipeline. First, raw PDFs were converted into structured Markdown using the open-source tool \emph{Marker} \cite{marker}. We then used rule-based methods to identify and isolate the dedicated EU Taxonomy disclosure section within each report.

A semi-automated workflow was employed to extract ground-truth data. We used an LLM (Gemini-Flash-2.0) with a zero-shot prompting strategy, followed by an LLM-based self-verification step, and concluded with a final human review. This process was applied to extract: (1) the list of declared economic activities, and (2) the quantitative numerical values for Taxonomy-eligible and Taxonomy-aligned Turnover, CapEx, and OpEx. General firmographic metadata (e.g., sector, country, revenue) was sourced from wikipedia and also human-verified. We show all the prompts in Appendix.

\subsection{Dataset Structure and Statistics}
For each of the 190 companies, the final dataset contains: (1) verified company metadata (country, number of employees, sector, industry, revenue); (2) the full annual report in machine-readable format (along with a version excluding the isolated Taxonomy section); (3) the extracted text of the Taxonomy disclosure section; (4) the human-verified list of reported economic activities (ground-truth labels); and (5) the human-verified quantitative KPIs (ground-truth values).

\paragraph{Dataset Composition}
The dataset is sectorally diverse, covering 11 distinct economic sectors, with \emph{Industrials}, \emph{Consumer Discretionary}, and \emph{Financials} being the most represented. It is, geographically concentrated, with companies based primarily in Germany (63\%) and Austria (20\%). Company size metrics (Table~\ref{tab:size_statistics}) are highly right-skewed, indicating the dataset is influenced by a number of very large, multinational corporations. This distribution is a direct result of the regulatory implementation timeline for the 2023 reporting cycle, mandatory disclosure was primarily limited to large public-interest entities under the scope of the Non-Financial Reporting Directive (NFRD)\cite{eu:nfrd:2014}.

\begin{table}[h!]
\small
\centering
\caption{Descriptive Statistics of Company Size Metrics $(N=190)$.}
\label{tab:size_statistics}
\begin{tabular}{lrr}
\toprule
\textbf{Statistic} & \textbf{Annual Revenue} & \textbf{Number of} \\
& \textbf{(in billion €)} & \textbf{Employees} \\
\midrule
Mean               & 45.18                     & 38436.82           \\
Median             & 2.78                      & 8982.00            \\
Std. Dev.          & 187.99                    & 101391.61          \\
Minimum            & 0.05                      & 49.00              \\
Maximum            & 1837.00                   & 684025.00          \\
\bottomrule
\end{tabular}
\end{table}

\paragraph{Reporting Statistics}
The analysis of reported data reveals significant variance. On average, companies reported 6.22 distinct economic activities (median 5.0, max 26), as shown in Table~\ref{tab:activity_statistics}. The most frequently declared activities relate to the buildings, transport, and energy sectors ).
\par
Crucially, the KPI statistics (Table~\ref{tab:kpi_percentage_statistics}) reveal a large gap between eligibility and alignment. While the median \emph{eligible} CapEx is 29.75\%, the median \emph{aligned} Turnover and OpEx are both 0.00\%, with aligned CapEx at only 0.42\%. This highlights that while many companies operate in eligible sectors, a very small fraction currently meet the strict Technical Screening Criteria for full alignment. Additional details can be found in Appendix \ref{sec:dataset}.

\begin{table}[h!]
\footnotesize
\centering
\caption{Descriptive Statistics of Reported Activities per Company (N=190)}
\label{tab:activity_statistics}
\begin{tabular}{lr}
\toprule
\textbf{Statistic}     & \textbf{Value} \\
\midrule
Mean                   & 6.22           \\
Median                 & 5.00           \\
Standard Deviation     & 4.63           \\
Minimum                & 1.00           \\
Maximum                & 26.00          \\
\bottomrule
\end{tabular}
\end{table}

\begin{table}[h!]
\footnotesize
\centering
\caption{Descriptive Statistics of EU Taxonomy KPIs (\%) $(N=190)$.}
\label{tab:kpi_percentage_statistics}
\begin{tabular}{@{}llrrr@{}}
\toprule
\textbf{KPI} & \textbf{Cat.} & \textbf{Mean} & \textbf{Med.} & \textbf{Std.}  \\
\midrule
\multirow{2}{*}{Turn.} & Elig. & 33.06 & 17.05 & 36.24   \\
                       & Alig. & 8.46 & 0.00 & 19.02   \\
\midrule
\multirow{2}{*}{CapEx} & Elig. & 39.44 & 29.75 & 34.93   \\
                       & Alig. & 13.20 & 0.42 & 24.69   \\
\midrule
\multirow{2}{*}{OpEx}  & Elig. & 29.84 & 10.37 & 36.33   \\
                       & Alig. & 9.23 & 0.00 & 20.85   \\
\bottomrule
\end{tabular}
\end{table}

\section{Methodology: Evaluating LLMs for Taxonomy Reporting}

\subsection{Models and Experimental Setup}
Our empirical evaluation utilizes five LLMs: Google's \texttt{Gemini 2.0 Flash} \cite{google2025gemini2_0_flash}, \texttt{Gemini 2.5 Flash} \cite{google2025gemini2_5_flash}, Meta's \texttt{Llama 4 Maverick 17B Instruct} \cite{meta2025llama4}, \texttt{gemma-3-12b-it} \cite{google2025gemma3}, and \texttt{Mistral-Small-3.1-24B-Instruct-2503} \cite{mistralai2025mistralsmall}. Models were accessed via Google Cloud Platform (GCP) Vertex AI or a high-performance computing (HPC) cluster. We list the prompts used in the Appendix \ref{sec:appendix_prompts}.

\subsection{Task 1: Multiclass Activity Prediction}
The objective of this task is to evaluate LLM performance in predicting a company's EU Taxonomy-relevant eligible economic activities, framed as a multi-label classification problem. Models must infer activities based on operational descriptions, not direct Taxonomy mentions. We tested three context variations to analyze the impact of informational depth: \begin{enumerate} \item \textbf{Prediction from Company Name:} The model uses only the company name, relying on its pre-existing knowledge. \item \textbf{Prediction from Metadata:} The model receives the company's name, industry sector, annual revenue, and employee count. \item \textbf{Prediction from Annual Report:} The model is given the full annual report text, with the dedicated EU Taxonomy disclosure section removed. \end{enumerate} Due to context window limitations, the 'Annual Report' variation was not run for \texttt{gemma-3-12b-it} and \texttt{Mistral-Small-3.1-24B-Instruct-2503}. Models were prompted to return a JSON list of identified activities, a justification, and a confidence score.

\subsection{Task 2: Binary Activity Classification}
To evaluate model precision at a more granular level, we test the ability of an LLM to validate whether a \emph{single} economic activity applies to a given company. A comprehensive approach would require over 28,000 classifications. To manage this cost, we used a representative sampling strategy, testing a subset of 30 companies (selected via stratified random sampling by sector) against the 116 activities actively reported by at least one company in our dataset.

\subsection{Task 3: Quantitative KPI Regression}
In this experiment we attempt to assess the ability of LLMs to perform regression by predicting the EU Taxonomy Key Performance Indicators (KPIs): Turnover, Capital Expenditure (CapEx), and Operational Expenditure (OpEx). The analysis focuses on the three \emph{eligible} KPIs, as the corresponding \emph{aligned} KPI data is heavily zero-inflated and unsuitable for this regression task. The task was performed by the Gemini 2.0 and 2.5 models under two contextual conditions: (1) \textbf{With Metadata} only, and (2) \textbf{With Report} (Taxonomy section removed).

\subsection{Task 4: Multi-Step Agentic Framework}
\begin{figure}[h]
\centering
\includegraphics[width=0.48\textwidth]{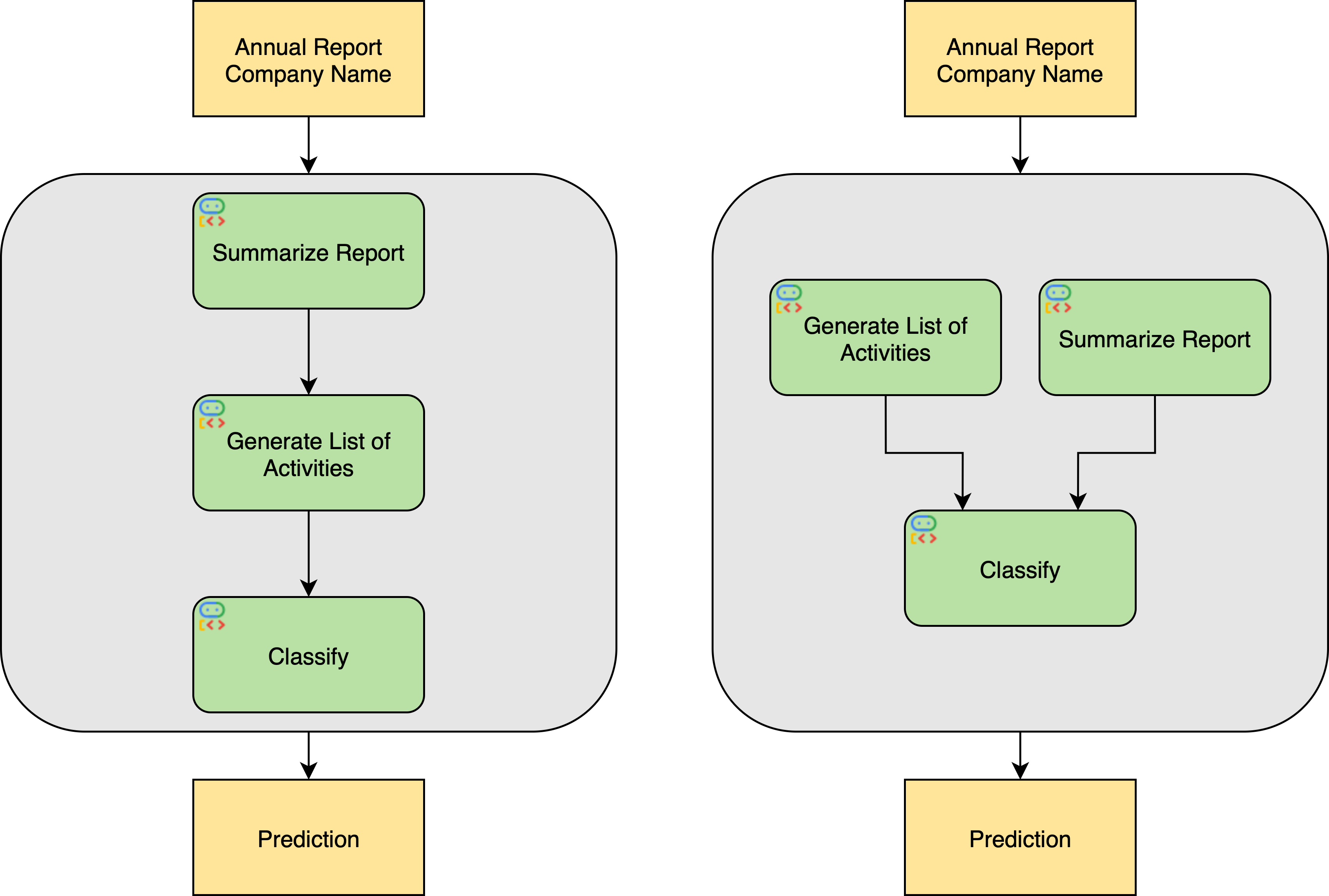}
\caption{Architectural diagrams of the two agentic workflows. Approach 1 (left) is a sequential pipeline. Approach 2 (right) uses a parallel structure to refine predictions.} \label{fig:agentic_workflows} \end{figure}

This experiment evaluates the effectiveness of multi-step, agentic workflows designed to simulate a more deliberative analytical process. The objective is to determine if a pipeline that combines candidate generation with a verification step can improve upon single-step prediction. We implemented and tested two distinct architectures, as illustrated in Figure~\ref{fig:agentic_workflows}. The implementation was done using Google's Agent Development Framework (ADK) \cite{google_adk_2025}.

\paragraph{Sequential Summarization and Classification}
This workflow follows a linear, three-step process: (1) The full annual report is processed by an LLM to generate a comprehensive summary. (2) This summary serves as the sole input for a multiclass prediction agent to generate an initial list of activities. (3) The list of candidates is passed to a binary classification agent for verification and refinement.

\paragraph{Parallel Generation with Summary-Informed Classification}
This workflow processes the full report and its summary in parallel: (1) The highest-recall model from Task 1 (Gemini Flash 2.5) generates a broad list of candidate activities directly from the full annual report. (2) In parallel, the report is summarized. (3) Finally, a binary classification agent validates the broad candidate list, using the summary as the contextual basis for its decision.

\section{Results and Analysis}
This section presents the empirical results of the four experiments, evaluating LLM performance on qualitative classification, quantitative regression, agentic workflows, and model reliability.

\subsection{Performance on Qualitative Activity Identification}
We first evaluated the models' ability to perform multi-label classification of economic activities (Experiment 1). As shown in Table~\ref{tab:exp1_full_results}, performance was moderate. \texttt{Gemini Flash 2.5} achieved the highest $F1$-score ($0.311$) when provided with concise company metadata. A clear trade-off emerged: models like \texttt{Gemini Flash 2.5} and \texttt{Mistral} favored high recall (up to $0.526$), while \texttt{Llama} achieved the highest precision ($0.348$), suggesting different models are suited for different workflow stages (e.g., discovery vs. verification).

\begin{table}[h!]
\centering
\caption{Performance Metrics for Multiclass Activity Prediction (Exp. 1). Best $F1$-score per model is bolded.}
\label{tab:exp1_full_results}
\resizebox{\columnwidth}{!}{%
\begin{tabular}{@{}llrrr@{}}
\toprule
\textbf{Model} & \textbf{Context} & \textbf{Precision} & \textbf{Recall} & \textbf{F1-Score} \\
\midrule
Gemini Flash 2.0 & Name         & \textbf{0.281} & \textbf{0.358} & \textbf{0.276} \\
                 & Company Info & 0.254 & 0.334 & 0.253 \\
                 & Report       & 0.241 & 0.332 & 0.245 \\
\addlinespace
Mistral          & Name         & 0.186 & \textbf{0.414} & 0.213 \\
                 & Company Info & \textbf{0.199} & 0.375 & \textbf{0.227} \\
\addlinespace
Llama            & Name         & 0.315 & 0.251 & 0.241 \\
                 & Company Info & \textbf{0.348} & \textbf{0.253} & \textbf{0.251} \\
                 & Report       & 0.203 & 0.172 & 0.162 \\
\addlinespace
Gemma            & Name              & \textbf{0.301} & \textbf{0.222} & \textbf{0.219} \\
                 & Company Info      & 0.250 & 0.217 & 0.203 \\
\addlinespace
Gemini Flash 2.5 & Name              & 0.268 & 0.452 & 0.297 \\
                 & Company Info      & \textbf{0.274} & 0.493 & \textbf{0.311} \\
                 & Report            & 0.213 & \textbf{0.526} & 0.275 \\
\bottomrule
\end{tabular}
}
\end{table}

In our granular binary classification task (Experiment 2), we also observed a context-dependent trade-off. Using only metadata, the model was highly precise ($67.86\%$) but had very low recall ($10.80\%$). Using the full report doubled the recall ($20.45\%$) but nearly halved the precision ($48.65\%$). This reinforces that concise context favors precision, while full-text context aids discovery.

\subsection{Failure on Quantitative KPI Prediction}
In contrast to the moderate success in classification, the models comprehensively failed at the quantitative KPI regression task (Experiment 3) in a zero-shot setting. As shown in Table~\ref{tab:kpi_regression_metrics}, the $R^2$ values were consistently negative across all predicted KPIs (e.g., $-0.2106$ for Turnover Eligible \%), indicating that the models' predictions were less accurate than a baseline model simply predicting the mean.

\begin{table}[h!]
\centering
\caption{Performance Metrics for KPI Regression Tasks (Exp. 3). The negative $R^2$ values indicate a complete failure to model the data.}
\label{tab:kpi_regression_metrics}
\resizebox{\columnwidth}{!}{%
\begin{tabular}{lrrr}
\toprule
\textbf{KPI Target} & \textbf{MAE} & \textbf{RMSE} & \textbf{$R^2$} \\
\midrule
\multicolumn{4}{l}{\textit{Percentage-Based KPIs}} \\
Turnover Eligible \% & 30.04 & 40.68 & -0.2106 \\
CapEx Eligible \% & 30.33 & 39.16 & -0.2856 \\
OpEx Eligible \% & 26.91 & 37.19 & -0.0020 \\
\midrule
\multicolumn{4}{l}{\textit{Absolute Value KPIs (in millions)}} \\
Turnover Eligible Value & 283,713 & 3,001,436 & -6192.88 \\
CapEx Eligible Value & 42,325 & 473,472 & -3845.34 \\
OpEx Eligible Value & 38,863 & 431,323 & -194.26 \\
\bottomrule
\end{tabular}
}
\end{table}

\begin{figure*}[t]
    \centering
    \includegraphics[width=\textwidth]{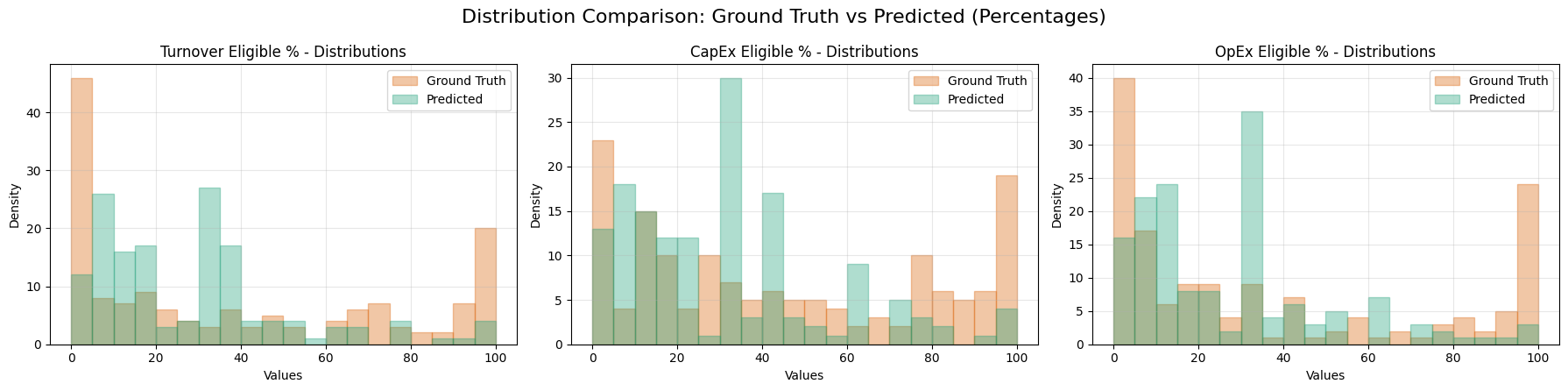}
    \caption{Comparison of ground-truth (orange) vs. predicted (green) KPI distributions. The model (gemini-flash-2.5) fails to capture the true data distributions.}
    \label{fig:kpi_distributions}
\end{figure*}

Error analysis confirmed this failure was systematic. Figure~\ref{fig:kpi_distributions} shows that the gemini-flash-2.5 predicted distributions (green) failed to replicate the complex, skewed, and multi-modal distributions of the ground-truth data (orange), often defaulting to a more uniform or unimodal output. The models are not grounding their numerical predictions in the text, demonstrating an inability to perform this specialized extraction and reasoning task without fine-tuning.

\subsection{Efficacy of Agentic Workflows}
Given the precision/recall trade-off in single-step classification, we evaluated a multi-step agentic framework (Experiment 4) to combine a high-recall generation step with a high-precision verification step. We tested two architectures on a sample of 50 companies using \texttt{Gemini Flash 2.5}.

\paragraph{Sequential Workflow} first summarized the report, then generated candidates from the summary, and finally verified them. This workflow underperformed the baseline, with a final $F1$-score of $0.2957$ (Table~\ref{tab:agentic_approach1_results}), likely due to information loss during the initial summarization.

\paragraph{Summarization Workflow} first generated a high-recall list of candidates from the \emph{full report} ($Recall=0.5404$) and, in parallel, generated a summary. This summary was then used as the context for a verification agent to filter the candidate list. As shown in Table~\ref{tab:agentic_approach2_results}, this approach was more effective. The verification step significantly improved precision (from $0.211$ to $0.3436$), resulting in a final $F1$-score of $0.3285$. This modestly outperforms the best single-step model ($0.311$) and demonstrates the value of a 'generate-then-verify' pipeline for improving reliability.

\begin{table}[h]
    \centering
    \caption{Performance of Sequential  Workflow.}
    \label{tab:agentic_approach1_results}
    \resizebox{\columnwidth}{!}{%
    \begin{tabular}{@{}lccc@{}}
        \toprule
        \textbf{Pipeline Stage} & \textbf{Precision} & \textbf{Recall} & \textbf{F1-Score} \\
        \midrule
        Step 1: Candidate Generation & 0.3487 & 0.3842 & 0.3250 \\
        \textbf{End-to-End (Full Pipeline)} & \textbf{0.3987} & \textbf{0.2961} & \textbf{0.2957} \\
        \bottomrule
    \end{tabular}
    }
\end{table}

\begin{table}[h]
    \centering
    \caption{Performance of Parallel  Workflow.}
    \label{tab:agentic_approach2_results}
    \resizebox{\columnwidth}{!}{%
    \begin{tabular}{@{}lccc@{}}
        \toprule
        \textbf{Pipeline Stage} & \textbf{Precision} & \textbf{Recall} & \textbf{F1-Score} \\
        \midrule
        Step 1: Candidate Generation & 0.2110 & 0.5404 & 0.2796 \\
        \textbf{End-to-End (Full Pipeline)} & \textbf{0.3436} & \textbf{0.4016} & \textbf{0.3285} \\
        \bottomrule
    \end{tabular}
    }
\end{table}

\subsection{Analysis of Influencing Factors: The Paradox of Context}
Our analysis of factors influencing multiclass prediction (Exp. 1) revealed a "paradox of context": providing more data is not always better. The average $F1$-score when using the full, unstructured annual report was $0.2264$, which was lower than using concise, structured company metadata ($0.2493$) or just the company name ($0.2492$). This suggests the extensive, noisy text of a full report impedes zero-shot inference for this task.

Other factors were also highly predictive. Company sector was a key determinant, with \emph{Utilities} and \emph{Real Estate} yielding the highest $F1$-scores ($0.4226$ and $0.3972$, respectively), while \emph{Financials} ($0.1551$) was the most challenging. A Random Forest analysis identified company size (specifically \texttt{log\_employees} and \texttt{log\_revenue}) as the most important predictive features of model performance, accounting for over 47\% of feature importance.

\begin{table}[h]
\centering
\caption{Top 5 Most Important Features from Random Forest Analysis}
\label{tab:rf_feature_importance}
\begin{tabular}{lr}
\hline
\textbf{Feature} & \textbf{Importance Score} \\ \hline
log\_employees & 0.2387 \\
log\_revenue & 0.2377 \\
sector\_Utilities & 0.0779 \\
sector\_Real Estate & 0.0687 \\
context\_with\_report & 0.0337 \\ \hline
\end{tabular}
\end{table}
\subsection{Analysis of Model Reliability: Confidence Calibration}
Finally, we assessed whether a model's self-reported confidence is a reliable indicator of its accuracy. For the multiclass classification task (Exp. 1), we found all models to be significantly miscalibrated. Table~\ref{tab:ece_results} shows high Expected Calibration Error (ECE)\cite{kadavath2022languagemodelsmostlyknow,tian2023justaskcalibrationstrategies} scores across the board.

Notably, \texttt{Gemini Flash 2.5}, the model with the highest recall, was also the most poorly calibrated, with its Expected Calibration Error (ECE) peaking at $0.684$ when using the full report. This indicates a strong tendency toward overconfidence; the model is not "aware" of its high rate of false positives. \texttt{Llama-4} was more stable, but still miscalibrated (ECE $\approx 0.40$). These findings show that in a zero-shot setting, verbalized confidence scores for this task are not dependable. 

\begin{table}[h!]
	\centering
    \caption{Expected Calibration Error (ECE) for Multiclass Prediction (Exp. 1) (Bin size $m = 15$). }
	\label{tab:ece_results}
	\resizebox{\columnwidth}{!}{%
	\begin{tabular}{llrr}
	\toprule
	\textbf{Model} & \textbf{Context Level} & \textbf{ECE Score} & \textbf{N} \\
	\midrule
	Gemini Flash 2.0 & Name Only & 0.392 & 1161 \\
	 & With Company Info & 0.361 & 1210 \\
	 & With Report & 0.353 & 1449 \\
	\midrule
	Gemini Flash 2.5 & Name Only & 0.600 & 1860 \\
	 & With Company Info & 0.611 & 2031 \\
	 & With Report & \textbf{0.684} & 2644 \\
	\midrule
	Gemma-3 & Name Only & 0.430 & 193 \\
	 & With Company Info & 0.483 & 602 \\
	\midrule
	Llama-4 & Name Only & 0.407 & 664 \\
	 & With Company Info & 0.390 & 622 \\
	 & With Report & 0.393 & 885 \\
	\midrule
	Mistral & Name Only & 0.606 & 3344 \\
	 & With Company Info & 0.546 & 2139 \\
	\bottomrule
	\end{tabular}%
}
\end{table}
\section{Discussion}
Our evaluation reveals a significant performance disparity between qualitative and quantitative tasks. For qualitative activity classification, the best single-step model achieved a moderate $F1$-score of $0.311$, which a multi-step agentic workflow modestly improving it to $0.3285$. This suggests current LLMs can act as "analyst assistants" for an initial discovery phase, especially high-recall models like \texttt{Gemini Flash 2.5}, but are not reliable for full automation. A human-in-the-loop is required to filter the many false positives.

In stark contrast, the models comprehensively failed at the quantitative KPI regression task, evidenced by consistently negative $R^2$ values. This highlights a critical gap between semantic text comprehension and the specialized numerical reasoning and extraction required to parse complex financial documents.

A key counterintuitive finding was the "paradox of context": providing the full, unstructured annual report often degraded classification performance compared to using concise, structured metadata. This suggests the extensive text of a full report introduces noise that impedes zero-shot inference \cite{wang2024limitssurveytechniquesextend}, and that system design should prioritize targeted, pre-processed information.

Finally, the models' self-reported confidence scores were found to be poorly calibrated, with ECE scores as high as $0.684$. We attribute this not just to model overconfidence, but to a fundamental information gap. The LLM acts as an external auditor, trying to replicate a complex internal assessment (the "ground truth") using only public data. It is being asked to make a context-rich decision with incomplete information, making its inability to produce reliable confidence scores an expected outcome. This reinforces findings from related work that domain-specific specialization is crucial \cite{birti2025optimizinglargelanguagemodels}.

\section{Conclusion}
This paper presented a systematic evaluation of LLMs on the core analytical tasks required for EU Taxonomy reporting, using a novel, curated dataset of 190 corporate reports which we have made public.

We make two primary contributions: (1) We release the first structured dataset for this multi-task (qualitative and quantitative) regulatory challenge. (2) We establish a comprehensive performance benchmark for prompting-based LLMs, demonstrating their current capabilities and limitations.

Our findings show that while LLMs fail at zero-shot quantitative KPI prediction, they show moderate promise as assistive tools for the qualitative task of activity identification. However, their outputs are not yet reliable for full automation. We conclude that the path toward robust automation will require specialized, fine-tuned models and sophisticated, human-in-the-loop agentic systems.

\section{Limitations}
Our findings are subject to three main limitations. First, our dataset is geographically concentrated, with a majority of firms from Germany ($63\%$) and Austria ($20\%$). Second, our study focuses on prompting-based strategies. The exclusion of fine-tuning as a methodology means our results represent a baseline for general-purpose models, not the potential of fully adapted expert models. Third, the rapid pace of AI development means these results are a snapshot in time; future models may overcome the limitations observed here.

\bibliography{custom}

\onecolumn
\section{Appendix}
\label{sec:appendix}

This appendix contains supplementary visualizations that provide additional detail on the dataset's characteristics.

\subsection{Dataset characteristics}
\label{sec:dataset}
\begin{figure}[h!]
    \centering
    \includegraphics[width=\textwidth]{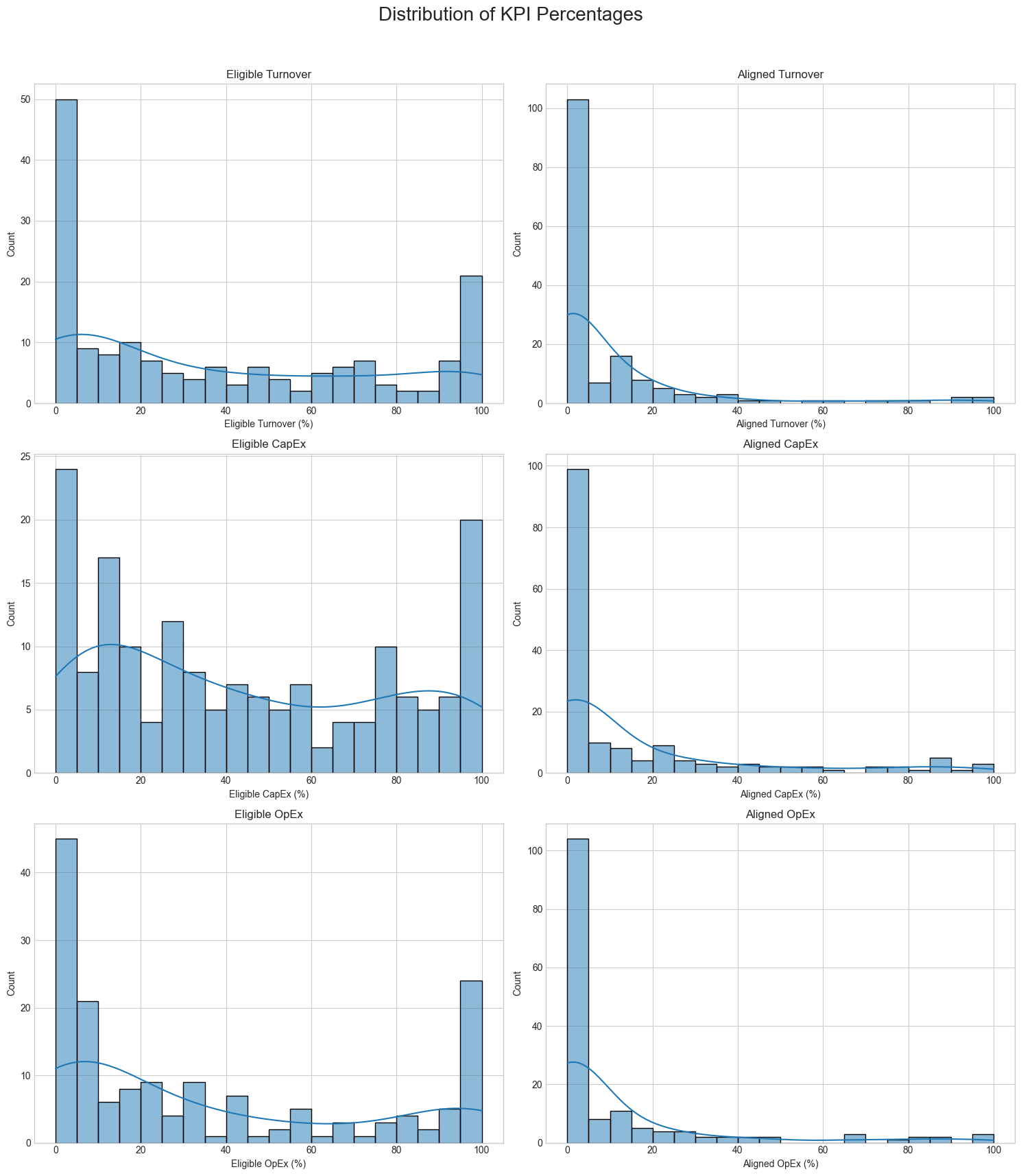}
    \caption{Detailed distributions of the six EU Taxonomy KPI percentages. The histograms for aligned KPIs (right column) show a strong concentration at or near zero, while eligible KPIs (left column) exhibit more varied distributions.}
    \label{fig:appendix_histograms}
\end{figure}

\clearpage

\begin{figure}[h!]
    \centering
    \includegraphics[width=\textwidth]{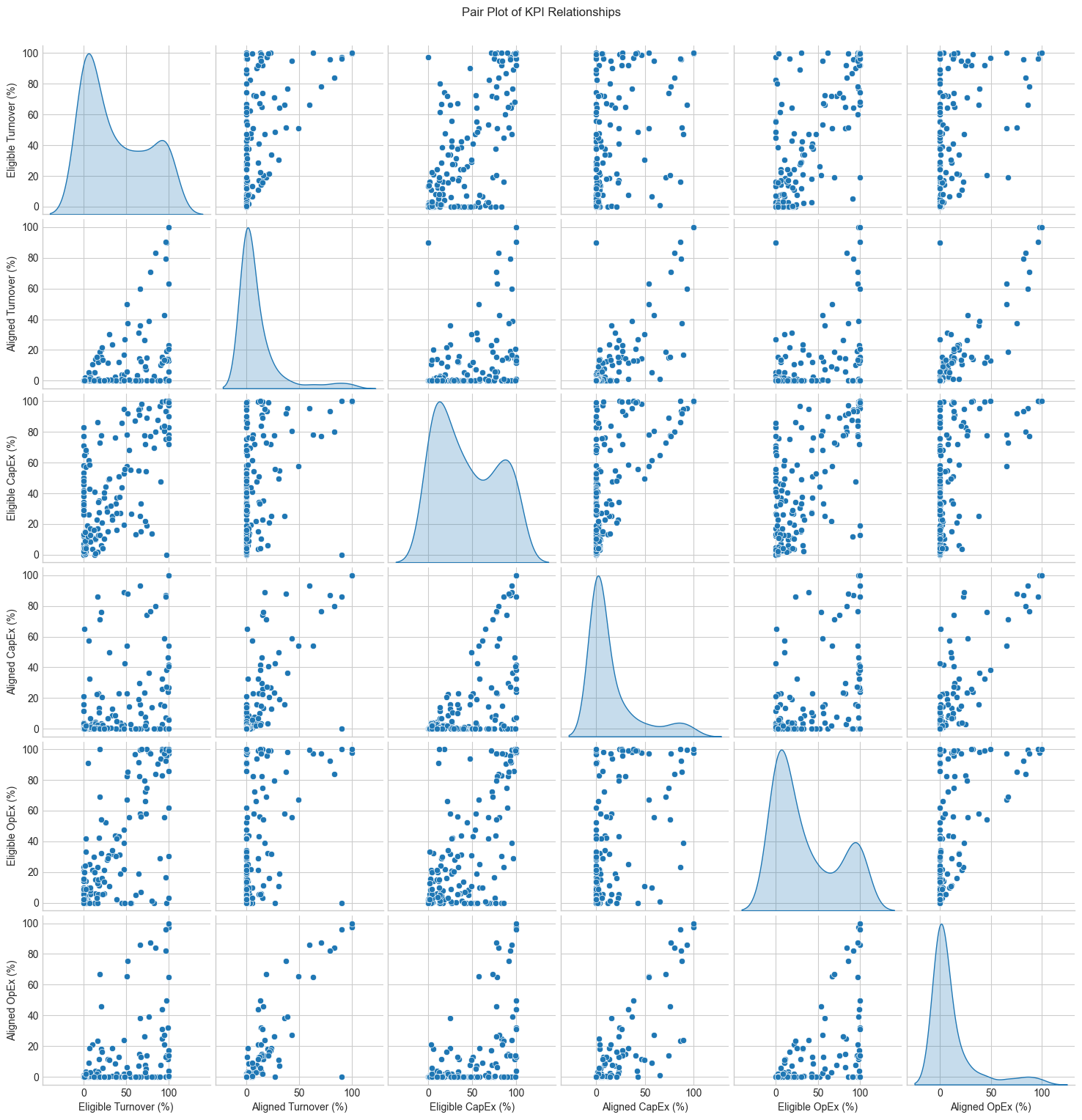}
    \caption{Pair plot matrix of the six EU Taxonomy KPI percentages. The diagonal shows the kernel density estimate (KDE) for each variable, while the off-diagonal plots show the scatter relationship between each pair of variables.}
    \label{fig:appendix_pairplot}
\end{figure}

\begin{figure}
    \centering
    \includegraphics[width=\textwidth]{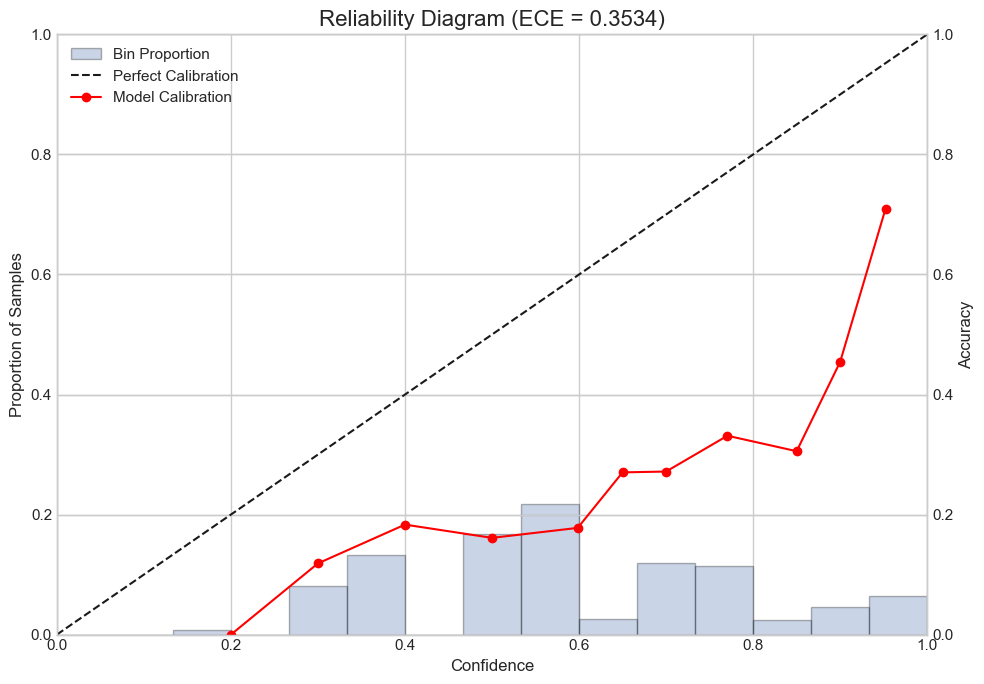}
    \caption{ECE Reliability Diagram Gemini Flash 2.0 Multiclass Task}
    \label{fig:ece}
\end{figure}

\begin{figure}
    \centering
    \includegraphics[width=\textwidth]{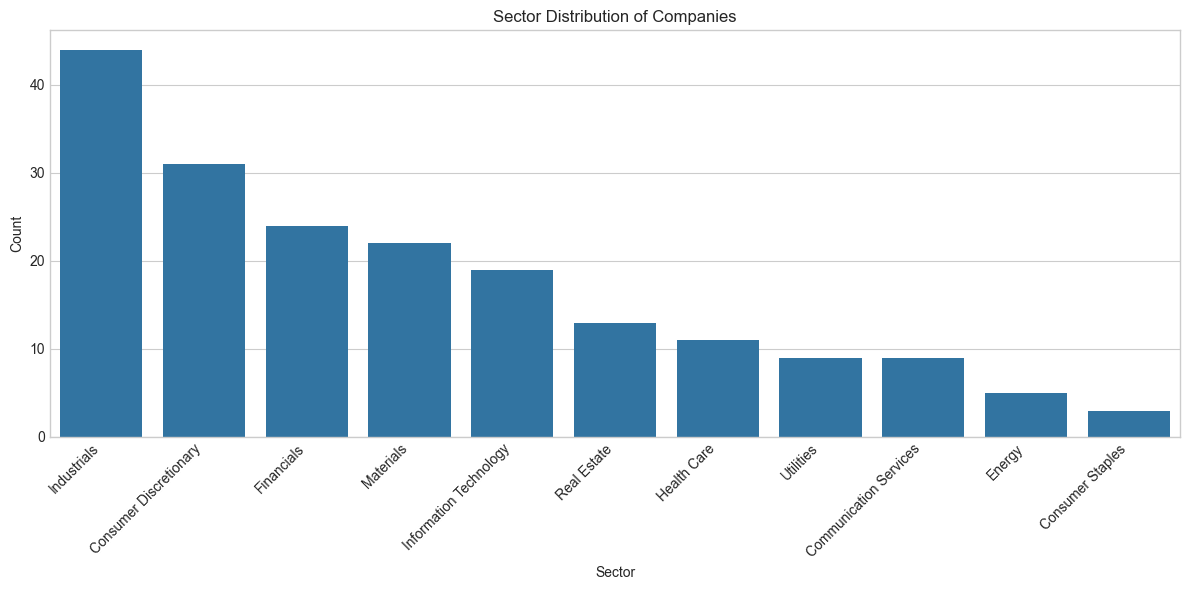}
    \caption{Sectoral distribution of companies in the dataset.}
    \label{fig:sector}
\end{figure}

\begin{figure}
    \centering
    \includegraphics[width=\textwidth]{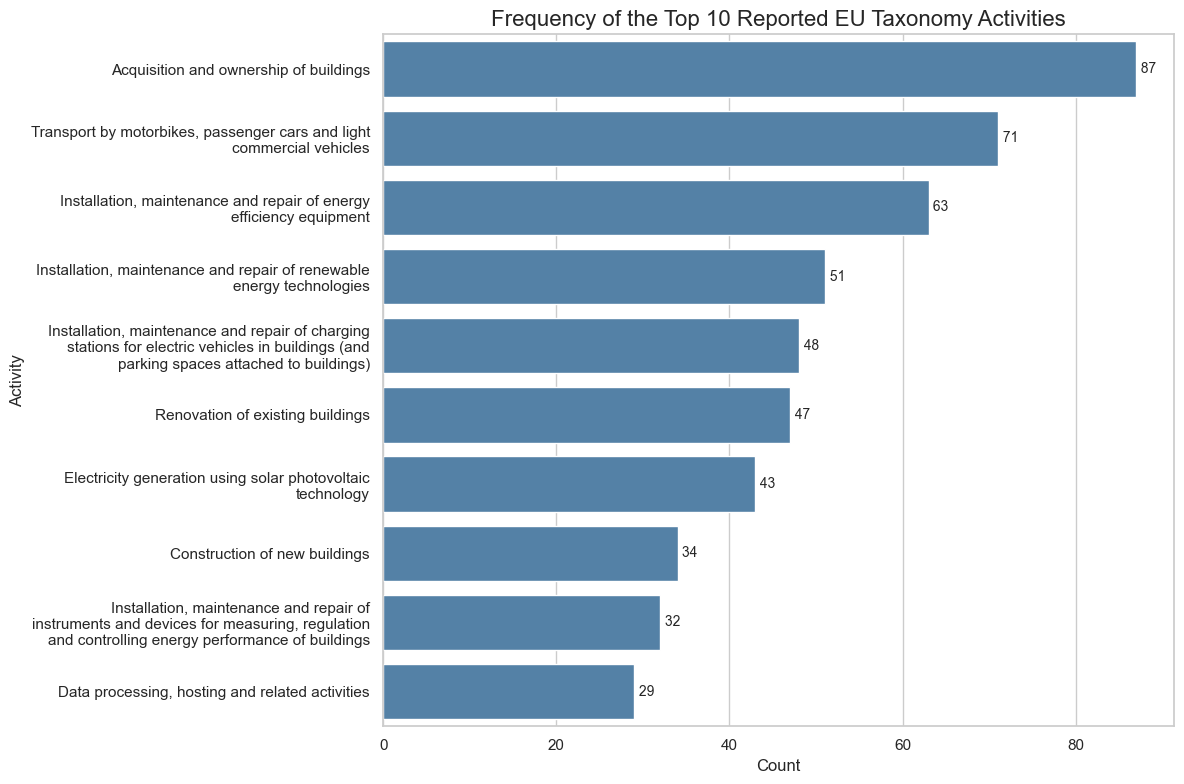}
    \caption{Frequency of the Top 10 Reported EU Taxonomy Activities}
    \label{fig:activities}
\end{figure}

\begin{figure}
    \centering
    \includegraphics[width=\textwidth]{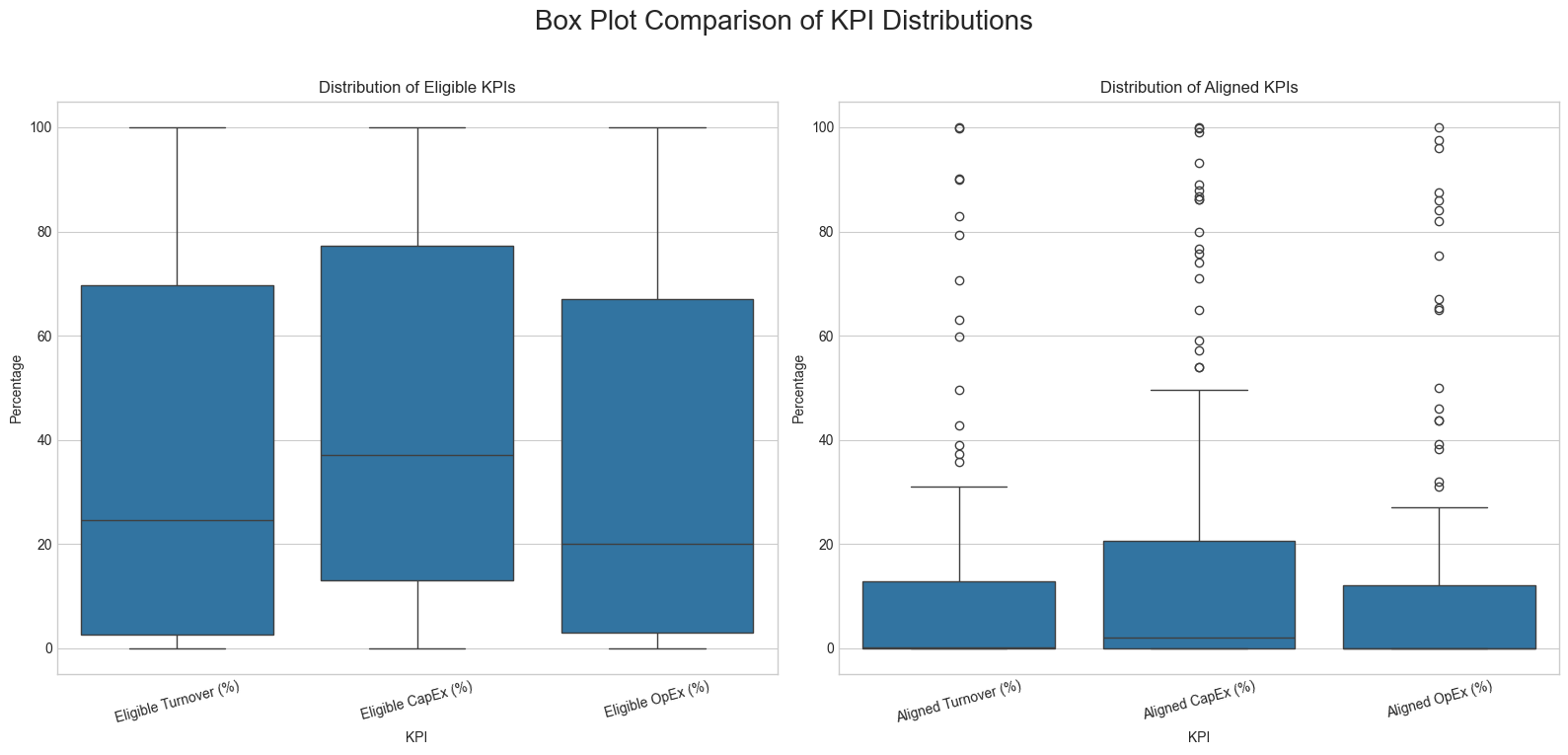}
    \caption{Box Plot Comparison of KPI Distributions.}
    \label{fig:kpi-box}
\end{figure}

\subsubsection{Prediction Plots}
\label{sec:prediction}
\begin{figure}
    \centering
    \includegraphics[width=\textwidth]{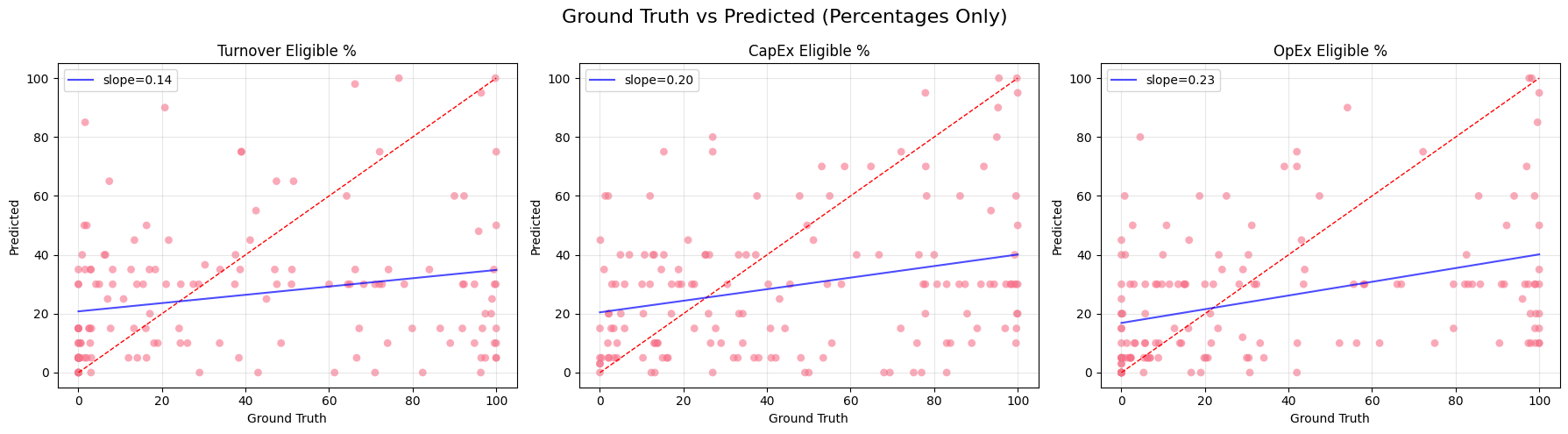}
    \caption{Comparison of ground-truth values versus model predictions for EU Taxonomy eligible KPIs.}
    \label{fig:kpi-scatter}
\end{figure}

\begin{figure}
    \centering
    \includegraphics[width=\textwidth]{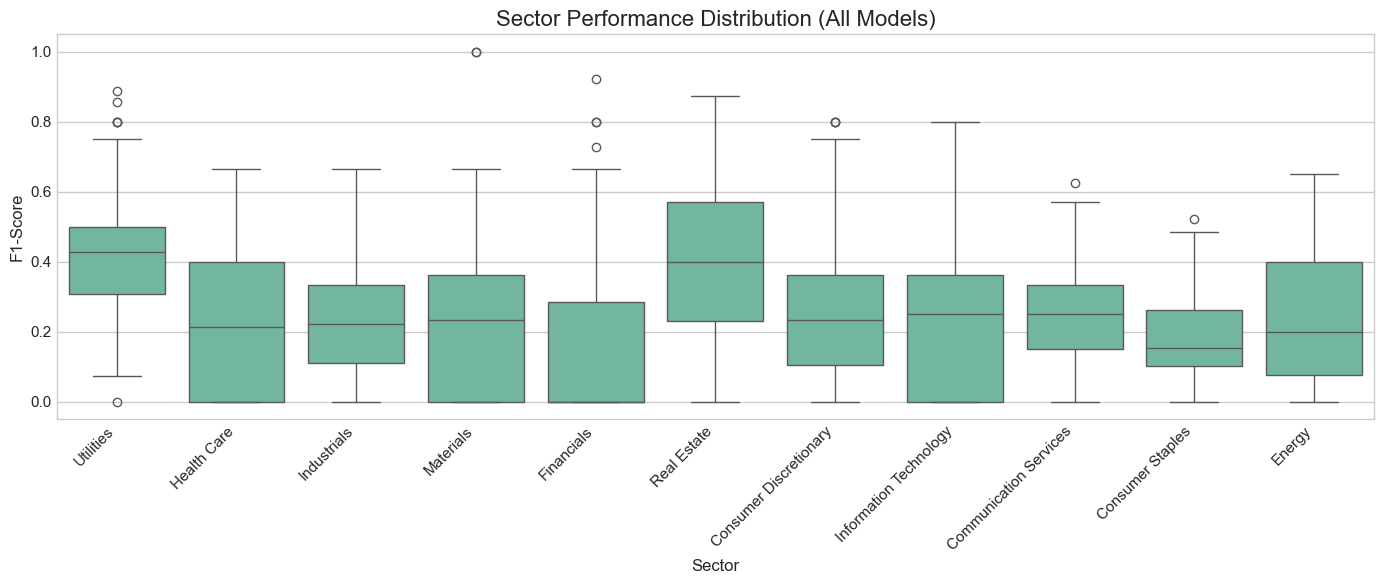}
    \caption{Distribution of F1-Scores by Company Sector across all models.}
    \label{fig:f1-sector}
\end{figure}

\begin{figure}
    \centering
    \includegraphics[width=\textwidth]{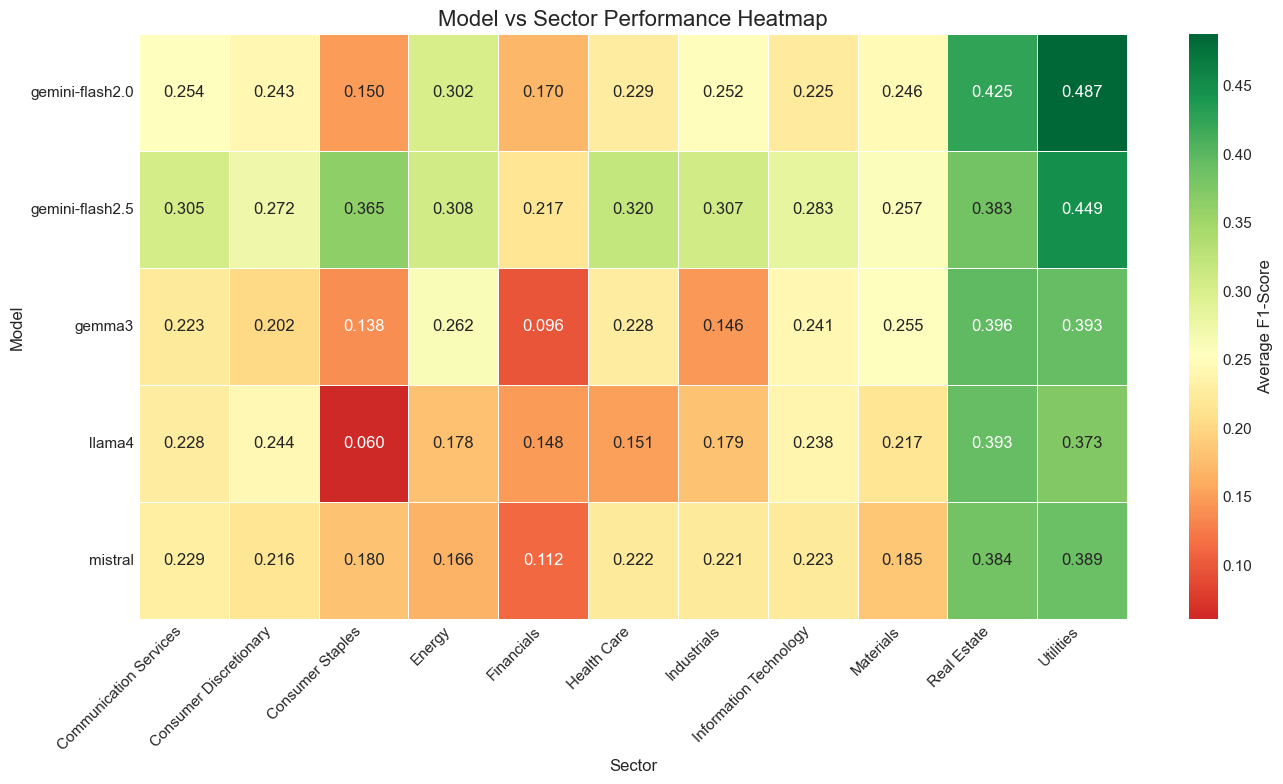}
    \caption{Heatmap of Average F1-Score by Model and Sector.}
    \label{fig:f1-heat}
\end{figure}

\clearpage

\subsection{Prompts}
\label{sec:appendix_prompts}
\subsubsection{Multiclass Prediction Prompt}
\label{sec:appendix_multi}
\begin{tcolorbox}[
  breakable, 
  colback=black!5!white, 
  colframe=black!75!black, 
  fontupper=\ttfamily, 
  title={Prompt designed for the multiclass prediction of EU Taxonomy activities (Experiment 1).},
  label={app:prompt:multiclass_prediction},
  arc=0mm 
]

**Task:** Your task is to act as a Sustainable Finance Analyst. Your goal is to thoroughly analyze the provided company `annual\_report` and predict all EU Taxonomy activities the company is likely involved in.

**Important Constraint:** The annual report will likely not mention the EU Taxonomy directly. Your task is to **infer** the activities based on the company's described operations, projects, products, services, and environmental initiatives.

**Context: The EU Taxonomy Framework**
The EU Taxonomy Regulation aims to help channel capital towards activities that substantially contribute to reaching the objectives of the European Green Deal. By providing clear, common definitions of environmentally sustainable economic activities, it helps to increase transparency and mitigate the risk of 'greenwashing'.

An economic activity must meet four conditions to qualify as environmentally sustainable:
1.  It must contribute substantially to one or more of the six environmental objectives.
2.  It must 'do no significant harm' (DNSH) to any of the other environmental objectives.
3.  It must be carried out in compliance with minimum social safeguards.
4.  It must comply with detailed technical screening criteria (TSC) established by the Commission.

The six environmental objectives are: climate change mitigation; climate change adaptation; sustainable use and protection of water and marine resources; transition to a circular economy; pollution prevention and control; and the protection and restoration of biodiversity and ecosystems.

**Core Directive:**
Your analysis must be comprehensive and evidence-based. It is better to include a plausible activity supported by textual evidence than to miss one.

**Inputs:**
* **Company Name:** \{company\_name\}
* **Company Information (JSON Object):**
    Company Information: \{company\_info\},
* **Annual Report:** The full text of the report will be at the End of the Instuctions
* **Authoritative List of EU Taxonomy Activities:**
   [
  "Afforestation",
  "Rehabilitation and restoration of forests, including reforestation and natural forest regeneration after an extreme event",
  "Forest management",
  "Conservation forestry",
  "Restoration of wetlands",
  "Conservation, including restoration, of habitats[1], ecosystems[2] and species",
  "Manufacture of renewable energy technologies",
  "Manufacture of equipment for the production and use of hydrogen",
  "Manufacture of low carbon technologies for transport",
  "Manufacture of batteries",
  "Manufacture of energy efficiency equipment for buildings",
  "Manufacture of other low carbon technologies",
  "Manufacture of cement",
  "Manufacture of aluminium",
  "Manufacture of iron and steel",
  "Manufacture of hydrogen",
  "Manufacture of carbon black",
  "Manufacture of soda ash",
  "Manufacture of chlorine",
  "Manufacture of organic basic chemicals",
  "Manufacture of anhydrous ammonia",
  "Manufacture of nitric acid",
  "Manufacture of plastics in primary form",
  "Manufacture of automotive and mobility components",
  "Manufacture of rail rolling stock constituents",
  "Manufacture, installation, and servicing of high, medium and low voltage electrical equipment for electrical transmission and distribution that result in or enable a substantial contribution to climate change mitigation",
  "Manufacturing of aircraft",
  "Manufacture, installation and associated services for leakage control technologies enabling leakage reduction and prevention in water supply systems",
  "Manufacture of plastic packaging goods",
  "Manufacture of electrical and electronic equipment",
  "Manufacture of active pharmaceutical ingredients (API) or active substances",
  "Manufacture of medicinal products",
  "Electricity generation using solar photovoltaic technology",
  "Electricity generation using concentrated solar power (CSP) technology",
  "Electricity generation from wind power",
  "Electricity generation from ocean energy technologies",
  "Electricity generation from hydropower",
  "Electricity generation from geothermal energy",
  "Electricity generation from renewable non-fossil gaseous and liquid fuels",
  "Electricity generation from bioenergy",
  "Transmission and distribution of electricity",
  "Storage of electricity",
  "Storage of thermal energy",
  "Storage of hydrogen",
  "Manufacture of biogas and biofuels for use in transport and of bioliquids",
  "Transmission and distribution networks for renewable and low-carbon gases",
  "District heating/cooling distribution",
  "Installation and operation of electric heat pumps",
  "Cogeneration of heat/cool and power from solar energy",
  "Cogeneration of heat/cool and power from geothermal energy",
  "Cogeneration of heat/cool and power from renewable non-fossil gaseous and liquid fuels",
  "Cogeneration of heat/cool and power from bioenergy",
  "Production of heat/cool from solar thermal heating",
  "Production of heat/cool from geothermal energy",
  "Production of heat/cool from renewable non-fossil gaseous and liquid fuels",
  "Production of heat/cool from bioenergy",
  "Production of heat/cool using waste heat",
  "Pre-commercial stages of advanced technologies to produce energy from nuclear processes with minimal waste from the fuel cycle",
  "Electricity generation from nuclear energy in existing installations",
  "Electricity generation from fossil gaseous fuels",
  "High-efficiency co-generation of heat/cool and power from fossil gaseous fuels",
  "Production of heat/cool from fossil gaseous fuels in an efficient district heating and cooling system",
  "Construction and safe operation of new nuclear power plants, for the generation of electricity and/or heat, including for hydrogen production, using best-available technologies",
  "Construction, extension and operation of water collection, treatment and supply systems",
  "Renewal of water collection, treatment and supply systems",
  "Construction, extension and operation of waste water collection and treatment",
  "Renewal of waste water collection and treatment",
  "Collection and transport of non-hazardous waste in source segregated fractions",
  "Anaerobic digestion of sewage sludge",
  "Anaerobic digestion of bio-waste",
  "Composting of bio-waste",
  "Material recovery from non-hazardous waste",
  "Landfill gas capture and utilisation",
  "Transport of CO2",
  "Underground permanent geological storage of CO2",
  "Desalination",
  "Water supply",
  "Urban Waste Water Treatment",
  "Sustainable urban drainage systems (SUDS)",
  "Phosphorus recovery from waste water",
  "Production of alternative water resources for purposes other than human consumption",
  "Collection and transport of non-hazardous and hazardous waste",
  "Treatment of hazardous waste",
  "Recovery of bio-waste by anaerobic digestion or composting",
  "Depollution and dismantling of end-of-life products",
  "Sorting and material recovery of non-hazardous waste",
  "Collection and transport of hazardous waste",
  "Remediation of legally non-conforming landfills and abandoned or illegal waste dumps",
  "Remediation of contaminated sites and areas",
  "Passenger interurban rail transport",
  "Freight rail transport",
  "Urban and suburban transport, road passenger transport",
  "Operation of personal mobility devices, cycle logistics",
  "Transport by motorbikes, passenger cars and light commercial vehicles",
  "Freight transport services by road",
  "Inland passenger water transport",
  "Inland freight water transport",
  "Retrofitting of inland water passenger and freight transport",
  "Sea and coastal freight water transport, vessels for port operations and auxiliary activities",
  "Sea and coastal passenger water transport",
  "Retrofitting of sea and coastal freight and passenger water transport",
  "Infrastructure for personal mobility, cycle logistics",
  "Infrastructure for rail transport",
  "Infrastructure enabling low-carbon road transport and public transport",
  "Infrastructure enabling low carbon water transport",
  "Low carbon airport infrastructure",
  "Leasing of aircraft",
  "Passenger and freight air transport",
  "Air transport ground handling operations",
  "Infrastructure enabling road transport and public transport",
  "Infrastructure for water transport",
  "Airport infrastructure",
  "Data processing, hosting and related activities",
  "Data-driven solutions for GHG emissions reductions",
  "Computer programming, consultancy and related activities",
  "Programming and broadcasting activities",
  "Software enabling physical climate risk management and adaptation",
  "Provision of IT/OT data-driven solutions for leakage reduction",
  "Provision of IT/OT data-driven solutions",
  "Close to market research, development and innovation",
  "Research, development and innovation for direct air capture of CO2",
  "Professional services related to energy performance of buildings",
  "Engineering activities and related technical consultancy dedicated to adaptation to climate change",
  "Consultancy for physical climate risk management and adaptation",
  "Non-life insurance: underwriting of climate-related perils",
  "Reinsurance",
  "Education",
  "Residential care activities",
  "Creative, arts and entertainment activities",
  "Libraries, archives, museums and cultural activities",
  "Motion picture, video and television programme production, sound recording and music publishing activities",
  "Construction of new buildings",
  "Renovation of existing buildings",
  "Installation, maintenance and repair of energy efficiency equipment",
  "Installation, maintenance and repair of charging stations for electric vehicles in buildings (and parking spaces attached to buildings)",
  "Installation, maintenance and repair of instruments and devices for measuring, regulation and controlling energy performance of buildings",
  "Installation, maintenance and repair of renewable energy technologies",
  "Acquisition and ownership of buildings",
  "Demolition and wrecking of buildings and other structures",
  "Maintenance of roads and motorways",
  "Use of concrete in civil engineering",
  "Emergency Services",
  "Flood risk prevention and protection infrastructure",
  "Nature-based solutions for flood and drought risk prevention and protection",
  "Repair, refurbishment and remanufacturing",
  "Sale of spare parts",
  "Preparation for re-use of end-of-life products and product components",
  "Sale of second-hand goods",
  "Product-as-a-service and other circular use- and result-oriented service models",
  "Marketplace for the trade of second-hand goods for reuse",
  "Hotels, holiday, camping grounds and similar accommodation"
]

**Instructions for Analysis and Prediction:**

1.  **Strategic Analysis of the Report:** Do not just read the report linearly. Analyze it strategically to find the most relevant information. Focus on sections such as:
    * 'Business Description', 'Our Operations', 'Products \& Services', or 'Business Segments'.
    * 'Sustainability Report' or 'Corporate Social Responsibility (CSR) Report'.
    * 'Management Discussion and Analysis (MD\&A)' for details on projects and investments.

2.  **Extract Core Business Activities:** From these sections, identify the company's fundamental business model. What does it physically produce, build, manage, or service?

3.  **Search for Thematic Evidence:** Look for descriptions of projects or processes that align with the themes of the six environmental objectives, even if the terminology is different. For example:
    * For **Circular Economy**, search for terms like "recycling," "waste reduction," "re-use," "circular design," "remanufacturing," "recycled content."
    * For **Pollution Prevention**, search for "emissions control," "waste treatment," "pollution reduction," "scrubbers," "hazardous substance management."
    * For **Water Protection**, look for "wastewater treatment," "water efficiency," "leakage reduction."

4.  **Map Evidence to Taxonomy Activities:** Based on the evidence you've gathered, match the company's operations to the specific activities in the `Authoritative List`.

5.  **Generate Structured Output:** Return your prediction as a JSON list of objects.

    * `activity\_name`: The exact name of the activity from the provided list.
    * `justification`: **This is critical.** You must justify your prediction by citing the specific evidence from the `annual\_report`. Briefly quote the relevant phrase or sentence that supports your conclusion.
    * `estimated\_relevance`: Assess how central the activity appears to be to the company's business model as described in the report. Use "High," "Medium," or "Low."
    * `confidence\_score`: A numerical score (0-1) reflecting your confidence that the described operation maps to the chosen Taxonomy activity.

**Example Output:**

[
  \{\{
    "activity\_name": "Manufacture of batteries",
    "justification": "The report states in the 'Strategic Investments' section: 'This year we began construction of our new giga-factory for the production of lithium-ion batteries for electric vehicles.'",
    "estimated\_relevance": "High",
    "confidence\_score": 0.98
  \}\}
  
]

Input:
Annual Report: \{annual\_report\}
Begin your analysis now.
\end{tcolorbox}

\subsubsection{Binary Prediction Prompt}
\label{sec:appendix_binary}
\begin{tcolorbox}[
  breakable, 
  colback=black!5!white, 
  colframe=black!75!black, 
  fontupper=\ttfamily, 
  title={Prompt designed for the binary classification of EU Taxonomy activities (Experiment 2).},
  label={app:prompt:binary_classification},
  arc=0mm 
]

**Task:** Your task is to act as a precision-focused Business Analyst. Your goal is to determine if a company performs a single, specific economic activity based on the provided company information and, when available, its annual report.

    **Core Directive:** Your analysis must be based **strictly** on the evidence within the provided texts. You must make a definitive binary decision (`true` or `false`). If there is no clear evidence that the company performs the activity, you must conclude `false`.

    **Context:** This classification is crucial for accurately categorizing a company's operations for regulatory, investment, and supply-chain analysis. The decision must be precise and directly supported by the provided information.

    ---

     **Inputs:**

    * **Company Information:** A JSON object or text describing the company's business, products, and services.
        * \{company\_info\}
    
    * **Activity to Classify:** The specific activity and its official description that you need to check for.
        * **Activity Name:** \{activity\_name\}
        * **Activity Description:** \{activity\_description\}
        * **Activity Sector:** \{activity\_sector\}
        
    * **Annual Report (Optional):** The text of the company's annual report. This input is optional and may not always be provided. It is provided at the end of this prompt.

    ---

     **Instructions for Analysis and Decision:**

    1.  **Understand the Target Activity:** First, carefully analyze the provided `activity\_name` and `activity\_description`. You must have a precise understanding of what the activity entails before searching for it.

    2.  **Scrutinize Company Data:** Thoroughly examine the `company\_info` and the `annual\_report` (if available). Search for explicit mentions or unambiguous descriptions of operations, products, services, projects, or revenue streams that directly match the `activity\_description`.

    3.  **Make a Binary Decision:** Based on your findings, decide if the company performs this activity.
        * **`true`**: Only if you find clear, direct evidence.
        * **`false`**: If there is no supporting evidence, if the evidence is ambiguous, or if it only relates to a minor or indirect part of the business (e.g., a supplier's activity).

    4.  **Determine Confidence:** Assign a confidence score between 0.0 and 1.0. This score reflects your certainty in the `applies` decision based on the quality of the evidence.
        * A direct quote mentioning the activity warrants a high score (e.g., 0.95-1.0).
        * A decision based on very strong inference from a product description would have a slightly lower score (e.g., 0.8-0.9).
        * A `false` decision due to a complete lack of evidence should have very high confidence (e.g., 0.99).

    5.  **Generate Structured Output:** Format your response as a **single JSON object**. Do not include any other text, explanation, or justification in your response. The JSON object must contain only two keys: `applies` (boolean), `confidence\_score` (float), and `justification` (string).

    ---

     **Example Scenario:**

    * **Company Info:** `\{\{"name": "Global Auto Corp", "description": "Global Auto Corp is a leading manufacturer of passenger cars, including a new line of fully electric vehicles (EVs)."\}\}`
    * **Activity Name:** `"Manufacture of low carbon technologies for transport"`
    * **Activity Description:** `"Manufacturing of transport equipment with zero or low tail-pipe CO2 emissions, such as electric vehicles."`

     **Correct Output for Example:**

    ```json
    \{\{
    "applies": true,
    "confidence\_score": 0.99,
    "justification": "The company explicitly states it manufactures fully electric vehicles, which directly aligns with the activity description."
    \}\}
    ```
    * **Annual Report (Optional):** The text of the company's annual report. This input is optional and may not always be provided. If it is provided, it should be used as the primary source of evidence.
        * \{annual\_report\}
  
    [Begin the Analysis based on the inputs]

\end{tcolorbox}

\subsubsection{KPI Prediction Prompt}
\label{sec:appendix_kpi}
\begin{tcolorbox}[
  breakable, 
  colback=black!5!white, 
  colframe=black!75!black, 
  fontupper=\ttfamily, 
  title={Prompt designed for prediction of EU Taxonomy KPIs (Experiment 3).},
  label={app:prompt:kpi_pred},
  arc=0mm 
]

You are an expert financial analyst with deep knowledge of the EU Taxonomy framework.

        The Challenge
        The provided annual report contains no direct mentions of the EU Taxonomy or its specific criteria. Your task is to analyze the company's general operations and financials and then use your internal, expert knowledge of the EU Taxonomy to predict the company's eligibility KPIs.

        Your Reasoning Process (Internal Monologue)
        1. First, read the report to form a high-level understanding of the company's business model: What does it produce? What services does it offer? Where do its revenues and capital expenditures come from?

        2. Next, compare this understanding of the company's activities against your internal knowledge of the broad economic sectors covered by the EU Taxonomy (e.g., energy, manufacturing, transport, buildings, etc.).

        3. Based on this comparison, estimate the percentage of the company's Turnover, CapEx, and OpEx that corresponds to these potentially eligible sectors.

        Your Final Task
        Provide your final quantitative prediction in the specified JSON format. Do not include your reasoning in the output; provide only the clean JSON object.

        Context:

        "company\_sector": \{company\_industry\},
        "financials": \{\{
            "currency": "EUR",
            "units": \{units\},
            "total\_turnover": \{total\_revenue\},
            "total\_capex": \{total\_capex\},
            "total\_opex": \{total\_opex\}
        \}\}
        
        The annual report:
        
        \{annual\_report\}

        Required Output JSON Format:

        \{\{
            "predictedTurnoverKPI": \{\{
                "eligiblePercentage": "number",
                "eligibleValue": "number",
                "alignedPercentage": "number",
                "alignedValue": "number"
            \}\},
            "predictedCapexKPI": \{\{
                "eligiblePercentage": "number",
                "eligibleValue": "number",
                "alignedPercentage": "number",
                "alignedValue": "number"
            \}\},
            "predictedOpexKPI": \{\{
                "eligiblePercentage": "number",
                "eligibleValue": "number",
                "alignedPercentage": "number",
                "alignedValue": "number"
            \}\}
        \}\}

\end{tcolorbox}

\subsubsection{Summarization Prompt}
\label{sec:appendix_kpi}
\begin{tcolorbox}[
  breakable, 
  colback=black!5!white, 
  colframe=black!75!black, 
  fontupper=\ttfamily, 
  title={Prompt designed for summarization of the annual report.},
  label={app:prompt:summarize},
  arc=0mm 
]

**Role:** You are an expert financial and sustainability analyst. Your specialization is in dissecting corporate annual reports to perform a detailed operational analysis and understand the core business of a company.

**Context:** You will be provided with the full text of a company's annual report. This report may not have a specific, pre-formatted section detailing all its economic activities in a standardized way.

**Primary Goal:** Your task is to perform a deep analysis of the annual report to produce an extensive and detailed summary. The summary must be structured into two main components: a comprehensive company profile and a detailed breakdown of the company's economic activities, which you will infer and construct from the report's content.

**Detailed Instructions:**

Carefully read and analyze the entire provided annual report. Pay close attention to sections such as "Business Description," "Strategy," "Management's Discussion and Analysis (MD\&A)," "Description of Operations," "Products and Services," "Segment Information," and any sustainability or ESG sections.

Based on your analysis, generate the output in two distinct parts as follows:

**Part 1: Comprehensive Company Profile**

Synthesize information from across the report to build a detailed company profile. This profile should be extensive and cover the following points in a clear, narrative format:

1.  **Company Overview:** State the company's full name, headquarters location, and the primary stock exchange it is listed on (if mentioned).
2.  **Mission and Vision:** Summarize the company's stated mission, vision, or core purpose.
3.  **Core Business \& Industry:** Describe the main industry the company operates in. What is its fundamental business model? How does it create value?
4.  **Products \& Services:** Detail the key products, services, or solutions the company offers. Group them logically if the company has multiple lines of business.
5.  **Operational Scale \& Geographic Footprint:** Describe the scale of the company's operations. Include details on key figures like revenue, number of employees, number of countries of operation, and key markets.
6.  **Major Business Segments:** If the company reports in segments, describe each segment's function, its contribution to the overall business (e.g., percentage of revenue), and the specific activities it undertakes.
7.  **Strategic Direction \& Outlook:** Summarize the company's forward-looking strategy, including stated goals for growth, innovation, market expansion, or strategic shifts as mentioned in the report.

**Part 2: Identification and Description of Economic Activities**

This is the most critical part of your task. Based on your holistic understanding from Part 1, you will identify, describe, and categorize the distinct economic activities the company engages in.

An "economic activity" is a process that uses a combination of resources (like labor, capital, goods, and services) to produce a specific set of products or services. A company may have one primary activity or several distinct ones, often corresponding to its business segments or major product lines.

For *each* distinct economic activity you identify, create a dedicated subsection with the following structure:

1.  **Economic Activity Title:** Create a concise, descriptive title for the activity. This title should be derived from the language and context of the report. Be as specific as the report allows (e.g., instead of a generic "Manufacturing," identify "Manufacturing of Consumer Electronics" or "Manufacturing of Heavy Industrial Machinery").
2.  **Detailed Description of the Activity:** Provide a comprehensive paragraph describing what this economic activity entails for the company. Explain the process, its purpose within the company's value chain, and its key characteristics.
3.  **Associated Business Segments:** List the official business segment(s) from the report that this activity falls under.
4.  **Key Products/Services Generated:** List the specific products or services that are the output of this economic activity.
5.  **Evidence-Based Justification:** This is essential. Provide a detailed justification for identifying this activity by quoting or directly paraphrasing specific sentences or passages from the annual report. Cite the section or page number if available. Explicitly connect the evidence from the report to your description of the activity.

If the company has only one primary economic activity, create a single, highly detailed section for it. If it has multiple, create a separate, structured section for each one.

**Output Format and Tone:**

*   The final output should be extensive, well-organized, and professional.
*   Use Markdown for clear headings (`\#`, `\#\#`, `\#\#\#`) and bullet points to structure your response.
*   Maintain an objective, analytical tone throughout the document.
*   Do not invent information. Your entire analysis must be grounded in the provided annual report.
*   A long and detailed response is highly encouraged to ensure a thorough summary.

---

**[START OF INPUT]**

**Annual Report:**
```
{report}
```

**[END OF INPUT]**
\end{tcolorbox}

\end{document}